\documentclass[letterpaper, 10 pt, conference]{ieeeconf}  

\IEEEoverridecommandlockouts                              

\usepackage{cite}
\usepackage{amsmath,amssymb,amsfonts}
\usepackage{algorithmic}
\usepackage{graphicx}
\usepackage{textcomp}

\usepackage[x11names]{xcolor}
\usepackage[group-separator={,},mode=text]{siunitx}

\usepackage[ruled,vlined]{algorithm2e}

\SetCommentSty{mycommfont}

\usepackage[labelformat=simple]{subcaption}
\DeclareCaptionLabelSeparator{periodspace}{.\quad}
\renewcommand{\thetable}{\arabic{table}}
\usepackage{booktabs}

\usepackage{array}
\newenvironment{conditions}
  {\par\vspace{\abovedisplayskip}\noindent\begin{tabular}{>{$}l<{$} @{${}={}$} l}}
  {\end{tabular}\par\vspace{\belowdisplayskip}}
  
\def\BibTeX{{\rm B\kern-.05em{\sc i\kern-.025em b}\kern-.08em
    T\kern-.1667em\lower.7ex\hbox{E}\kern-.125emX\}}}

\makeatletter
\let\NAT@parse\undefined
\makeatother
\usepackage{hyperref}
\hypersetup{
  colorlinks=true,
  citecolor=SpringGreen4
}
\usepackage[nameinlink]{cleveref}  
\usepackage{orcidlink} 

\title{\LARGE \bf
OdoViz: A 3D Odometry Visualization and Processing Tool
}

\author{Saravanabalagi Ramachandran* \orcidlink{0000-0001-6543-5345} and John McDonald* \orcidlink{0000-0001-9225-673X}
\thanks{*Saravanabalagi Ramachandran and John McDonald are with Lero - the Irish Software Research Centre and the Department of Computer Science, Maynooth University, Maynooth, Ireland
        {\tt\{saravanabalagi.ramachandran, john.mcdonald\}@mu.ie}}
}

\begin{document}

\maketitle
\thispagestyle{empty}
\pagestyle{empty}

\begin{abstract}
OdoViz is a reactive web-based tool for 3D visualization and processing of autonomous vehicle datasets designed to support common tasks in visual place recognition research. The system includes functionality for loading, inspecting, visualizing, and processing GPS/INS poses, point clouds and camera images. It supports a number of commonly used driving datasets and can be adapted to load custom datasets with minimal effort. 
OdoViz's design consists of a \textit{slim server} to serve the datasets coupled with a \textit{rich client} frontend. This design supports multiple deployment configurations including single user stand-alone installations, research group installations serving datasets internally across a lab, or publicly accessible web-frontends for providing online interfaces for exploring and interacting with datasets.
The tool allows viewing complete vehicle trajectories traversed at multiple different time periods simultaneously, facilitating tasks such as sub-sampling, comparing and finding pose correspondences both across and within sequences. This significantly reduces the effort required in creating subsets of data from existing datasets for machine learning tasks. Further to the above, the system also supports adding custom extensions and plugins to extend the capabilities of the software for other potential data management, visualization and processing tasks. The platform has been open-sourced to promote its use and encourage further contributions from the research community.
\end{abstract}

\section{Introduction}

Although autonomous vehicles research can be traced back to the early 1980's \cite{dynamic_vision_2007} \cite{1986_the_man} \cite{highlights_of_robot_cars_history}, the past decade has witnessed dramatic progress in the field. A core ingredient of this progress has been the use of data-driven, and in particular, deep learning techniques. In order for these approaches to be possible, multiple research groups and companies have led significant efforts to collect and release large-scale annotated datasets. These datasets facilitate the training of new models and approaches, whilst also providing a means of tracking and benchmarking progress on various research challenges within the field. Probably most notable here is the KITTI dataset \cite{Kitti_Geiger2012CVPR} from Karlsruhe. However, many others have also been released, often targetting more challenging conditions, particular sensor setups, specific problems, etc. Examples include: St Lucia \cite{St_Lucia_Warren2010}, CMU Seasons \cite{badino_2012_real_time_topometric_localization_cmu_cvg}, Oxford Robotcar \cite{RobotCarDatasetIJRR}, BDD100K \cite{yu2018bdd100k}, Woodscape \cite{yogamani2019_woodscape}, etc.

\begin{figure}
    \begin{subfigure}{0.23\columnwidth}
        \centering
        \includegraphics[width=\columnwidth]{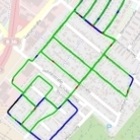}
        \caption{KITTI}
        \label{fig:dataset_kitti}
    \end{subfigure}
    \begin{subfigure}{0.23\columnwidth}
        \centering
        \includegraphics[width=\columnwidth]{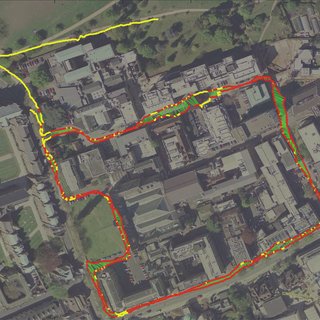}
        \caption{New College}
        \label{fig:dataset_new_college}
    \end{subfigure}
    \begin{subfigure}{0.23\columnwidth}
        \centering
        \includegraphics[width=\columnwidth]{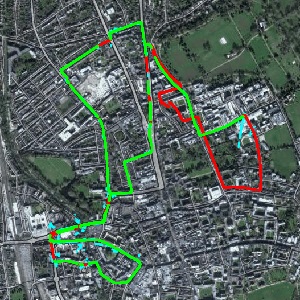}
        \caption{Robotcar}
        \label{fig:dataset_robotcar}
    \end{subfigure}
    \begin{subfigure}{0.25\columnwidth}
        \centering
        \includegraphics[width=\columnwidth]{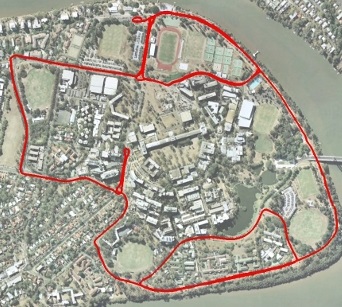}
        \caption{St Lucia}
        \label{fig:dataset_lucia}
    \end{subfigure}
    \caption{Path overlaid satellite images provided by different popular datasets}
    \label{fig:dataset}
\end{figure}

Typically the first step in determining a dataset's applicability for a given project is to visualize and analyze its included trajectories. For example, assessing a dataset's suitability for visual place recognition research requires analysing the geographical extent and degree of overlap between individual trajectories. Many popular public driving datasets only provide a single view of the included trajectories overlaid on a static aerial or satellite image (see \autoref{fig:dataset}). More recently datasets such as BDD100K have included executable scripts to visualize the top view of the individual trajectories. However, such scripts typically do not include functionality for loading multiple trajectories, analyzing the overlaps, and inspecting individual pose data (e.g. GPS, heading, image, and other sensor data).

\begin{table}
\centering
\begin{tabular}{ccc} \toprule
Dataset & Pose File Format & Camera Images \\\toprule
New College \cite{New_college_CumminsNewmanIJRR08} & TXT & JPG \\
KITTI Odometry \cite{Kitti_Geiger2012CVPR} & TXT & PNG \\
CMU Seasons \cite{badino_2012_real_time_topometric_localization_cmu_cvg} & NVM & JPG \\
St Lucia \cite{St_Lucia_Warren2010} & LOG & Bayer PNG \\
Oxford Robotcar \cite{RobotCarDatasetIJRR} & CSV & Bayer PNG \\
BDD100K \cite{yu2018bdd100k} & JSON & MOV Video \\\bottomrule
\end{tabular}
\caption{Different file formats used by popular driving datasets}
\label{tab:datasets_file_formats}
\end{table}

Given a suitable dataset, utilising it requires the researcher to become familiar with its organisation and structure so as to curate suitable data for training and evaluating new models. Many file formats have been adopted by different datasets (as shown in \autoref{tab:datasets_file_formats}), with each dataset having its own sensor types, positioning, and configuration. Ideally, software development kits (SDKs) are released alongside the datasets to simplify their use (e.g. Robotcar SDK \cite{robotcar_sdk}, BDD100K Toolkit \cite{bdd100k_toolkit}, etc.). However, in our experience there is still a steep learning curve associated with most driving datasets. In particular, one has to expend a lot of effort and time to setup and become familiar with the SDK. Furthermore, given the variations between the APIs, the use of different programming languages, etc., any code developed to perform additional tasks will have limited portability across datasets. 

Creating training sets from one or more datasets usually involves (i) sampling or selecting the images required, (ii) collecting corresponding information regarding the vehicle's pose from the Global Positioning System (GPS) and Inertial Navigation System (INS) data, (iii) applying positional and rotational offsets, and (iv) finding pose correspondences within or across the trajectories. For many computer vision challenges, it is often preferable to select images of importance based on their location on the map e.g. choosing images of landmarks, images captured at intersections, or images of the same place captured from different viewpoints. 
The SDKs are often limited to only parsing the pose files to obtain GPS/INS data for images and converting the images to standard file formats and do not include  an interactive interface to perform subsampling or finding such correspondences.

In this paper, we present OdoViz, a new 3D odometry visualization and processing tool to solve many of these odometry visualization and processing challenges using a single extensible platform. OdoViz is web-based, flexible, extensible, easy-to-use, and supports common odometry file formats with customizable scene and offset settings. The system allows the user to perform operations such as sampling, comparing and finding pose correspondences within and across multiple trajectories. It also allows loading, inspecting, visualizing and processing GPS/INS poses, point clouds and camera images. OdoViz has out of the box support for popular driving datasets, including: Oxford Robotcar \cite{RobotCarDatasetIJRR}, CMU Seasons \cite{badino_2012_real_time_topometric_localization_cmu_cvg}, BDD100K \cite{yu2018bdd100k}, St. Lucia \cite{St_Lucia_Warren2010}, New College \cite{New_college_CumminsNewmanIJRR08} and KITTI \cite{Kitti_Geiger2012CVPR}; and can be simply extended to support custom datasets. We further include plugins for importing and exporting settings, and extensions for additional tasks, (i) analyzing top-k matches in an image retrieval benchmark of a feature extractor, and (ii) visualizing topological nodes on a loaded trajectory. The system also supports adding custom extensions and plugins for additional tasks. 



We explain the design and architecture of OdoViz in \Cref{section:design}, break down each of the core modules in \Cref{section:core_modules}, elaborate sampling and finding correspondences functionalities in \Cref{section:functionality}, and discuss the extensions and plugins in \Cref{section:extensions_and_plugins}.

\section{Related Work}
Processing of autonomous driving datasets typically involves either directly accessing the data through hand developed code, utilising bespoke SDKs designed for individual datasets, using related tools from fields such as photogrammetry, or employing dedicated autonomous vehicle dataset frameworks.   

Probably the most notable tool from the field of photogrammetry for visualization of driving datasets is Visual SFM \cite{visual_sfm}. This tool allows loading images, performing Structure from Motion (SfM), saving and loading \textit{.nvm} files, and viewing the resultant point cloud and image associated with each camera pose. Being a full fledged SfM tool, its support does not extend beyond \textit{.nvm} files and therefore does not support many of the additional file formats included with vehicle datasets. 

More recently a number of companies within the autonomous vehicles space have released tools designed specifically to address some of these issues and to better support research in the space.  

Webviz \cite{webviz} is a web-based tool developed by Cruise for general robotics data inspection. The tool provides visual insights on ROS bag files and allows connecting to a live robot or simulation. Webviz allows custom data visualization layouts from a collection of configurable panels for displaying information like text logs, 2D charts and 3D depictions of the vehicle's environment.

Autonomous Visualization System \cite{avs_auto} (AVS) is an open and modular web tool developed by Uber to load and navigate though individual poses of a trajectory. It also allows inspecting GPS/INS and point clouds on a frame by frame basis. However, it does not allow viewing of complete trajectories. 


We also note recent additions to the popular Open3D library \cite{Open3d_Zhou2018}, in particular the Open3D-ML extension which extends the library to support common machine learning tasks in autonomous vehicles research. These extensions are targeted at 3D deep learning tasks such as 3D object detection and semantic segmentation.

The above tools address and important need within the community by providing rich frameworks for processing and visualisation at the local level of the vehicle, i.e. targeting egocentric tasks such as real-time visualisation and playback of vehicle sensor data, 3D object detection, etc. 
However, they are not directly applicable in more global level tasks such as the use-cases considered in this paper e.g. identifying corresponding poses within or across trajectories, or for visualizing loop closures.

To the best of our knowledge OdoViz is unique in this regard in that it is the only tool of which we are aware that supports loading, viewing and processing of complete trajectories, 
and performing common odometry tasks such as sampling and finding pose correspondences. OdoViz was initially created as a small in-house tool to load and inspect different datasets and to reduce the effort required to incorporate new datasets within our research. The tool has been in development since 2019, with multiple features added to the software since then to aid our research. At this point it has grown mature to act as generic extensible 3D odometry visualization and dataset curation tool, we have open-sourced\footnote{Source-code is available at \url{https://github.com/robotvisionmu/odoviz}} the work under the MIT licence for the benefit of the wider research community. 

A live instance\footnote{Live instance is available at \url{https://odoviz.cs.nuim.ie}} of OdoViz is hosted online for preview purposes. Documentation and a number of video tutorials\footnote{Video Tutorials are available at \url{https://www.youtube.com/playlist?list=PLKIavzsN4tuGi1SKDSPss0M8v4zswVEn9}} have been made available to assist the user in getting to know the system and completing common tasks. Documentation on extending the system to support a new dataset is also provided.
\section{Design}
\label{section:design}

The OdoViz architecture consists of (i) a front-end providing a \textit{rich client} built on React, Redux, and Redux Saga, and (ii) a backend server designed to act as a JSON API based \textit{thin server} primarily for serving files. The reactive front-end equipped with a threejs 3D environment, provides a sophisticated user interface. The complete application is loaded as a Single Page Application (SPA), ensuring full functionality and minimum processing latency after the application is loaded into memory. Internet connectivity is however required to load new files from the server.

The system is designed around the following principles: 
\begin{itemize}
\item{\textbf{Web-based:}} The software is web based and therefore runs without compilation in all major operating systems directly in the browser, and adding new and debugging existing features easily with hot-reloading. Dependency resolution is performed automatically using NPM (Node Package Manager) where webpack links the dependencies, minifies and bundles the app in a JavaScript file which can then be served along with a static HTML file. Web-based tools further allow providing a rich, modern, reactive and easily tweakable user-interface while being highly accessible and portable.

\item{\textbf{Extensible:}} The system is designed with extensibility in mind. This is achieved by associating a single parser file with each dataset that includes paths to its various file locations, along with information used for extracting data from the info file and linking the data with the images.
Hence, adding support for a new dataset requires only adding a new parser file. This facilitates the handling of multiple datasets and file formats within a single platform. OdoViz can also be equipped with extensions that allow inspecting and modifying loaded odometry data, reading external files, comparing images, etc.

\item{\textbf{Real-time:}} Data is processed and visualized in real-time. Powered by JavaScript and React/Redux, OdoViz has asynchronous execution and immutablility at its core i.e. it runs tasks asynchronously without blocking the UI, updates the data in an immutable fashion by updating existing pointers to point to the new data, and displays the changes reactively. This pipeline allows the changes to variables to reflect on the visualization near instantly.
\end{itemize}

\begin{figure}
    \includegraphics[width=\columnwidth]{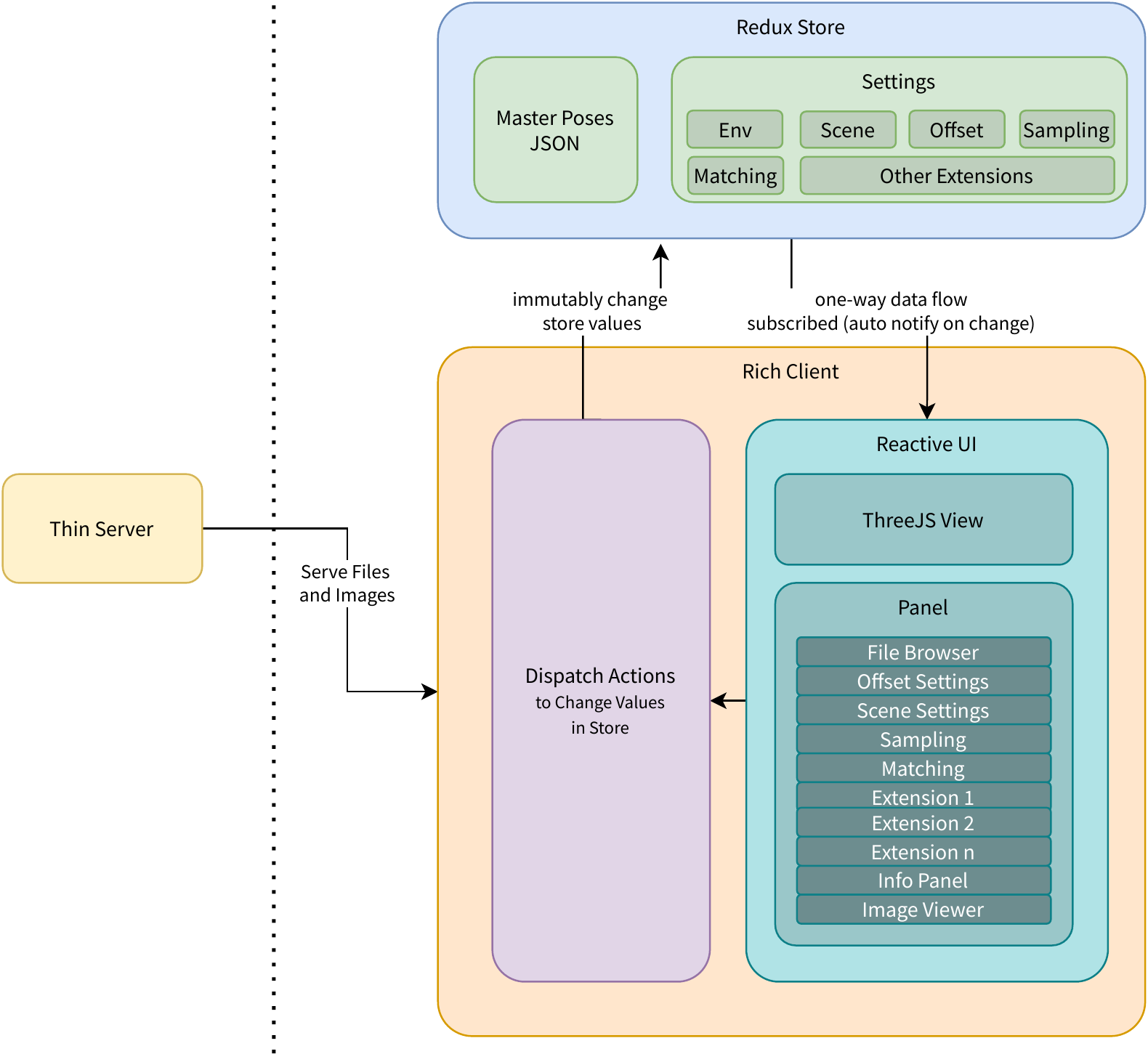}
    \caption{Overall architecture of OdoViz}
    \label{fig:arch}
\end{figure}
\autoref{fig:arch} shows the architecture of OdoViz including both the \textit{thin-server} and \textit{rich-client} design. The front-end has a redux \textit{store} to manage the app's state in a nested object literal which acts as a single source of truth for all of the app's data and configuration. The \textit{store} is read-only and can only be immutably modified as the user interacts with the app using \textit{reducers}. Reducers are pure functions that take in the previous state and the current action to produce a new state. The reactive user-interface comprises of a collection of components, each of which subscribes to required store data and populates its template. As data changes in the store, all subscribed components re-render to reflect the updates. This data flow allows the app to be more predictable, traceable and reproducible.

The OdoViz architecture allows the system to be deployed in a number of different ways, including:
\begin{itemize}
    \item a browser-based front-end tool consuming data served on the same computer using a NodeJS server.
    \item a standalone single user installation software packaged up individually for various operating systems using a packager system such as \href{https://www.electronjs.org/}{ElectronJS}.
    \item an in-house app in an organization or research group with a common server to serve files stored centrally, which can then be consumed by its members.
    \item an online interactive visualization tool for publicly available datasets that will allow the users to visually explore and inspect the various trajectories before downloading them. With all the computation taking place in the client, this allows setting up a server with virtually no compute load, hosted only to serve files. For servers with limited network bandwidth, a smaller subset of the dataset can be made available to be interactively viewed. The software can additionally be customized to have only limited features and can also be setup to allow users to selectively download desired trajectories.
\end{itemize}

\section{Core Modules}
\label{section:core_modules}
In this section, details are provided for each of the core modules that make up the overall system.

\subsection{Data Parser}
\label{sec:data_parser}
OdoViz supports loading Oxford Robotcar GPS, INS and Odometry CSV files, BDD100K info JSON files, CMU Seasons NVM files, KITTI \textit{txt} files, New College \textit{txt} files and St. Lucia \textit{log} files. OdoViz also supports loading other generic \textit{.nvm} bundler files with point-cloud visualization. Additional parser files can be easily added to provide support for custom datasets. Data parsed from any source is consistently loaded into the browser's memory in a master JavaScript object literal containing the following keys: \textit{index}, \textit{timestamp}, \textit{position}, \textit{orientation}, \textit{gps}, \textit{altitude}, \textit{imageIndex}, and \textit{image}. Visualization, sampling, matching and all other tasks can then use this data for processing. In cases where it may not be possible to load proprietary file formats, the files must first be converted to any \textit{open format} and then parsed accordingly. For example, \textit{.mat} files, which are binary MATLAB files that store workspace variables, should be loaded in MATLAB and then exported as one or more \textit{.csv} files.

\begin{figure}
    \begin{subfigure}{0.49\columnwidth}
        \centering
        \includegraphics[width=\columnwidth]{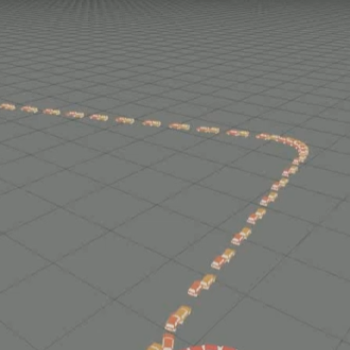}
        \caption{}
    \end{subfigure}
    \begin{subfigure}{0.49\columnwidth}
        \centering
        \includegraphics[width=\columnwidth]{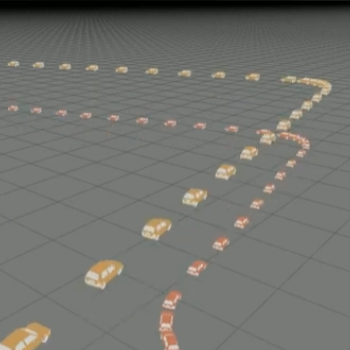}
        \caption{}
    \end{subfigure}
    \caption{Screenshots showing the difference between flattened and time-encoded z-axis. (a) Flat z-axis, only colored based on index of the pose (b) Time-encoded z-axis, z-offset added based on the index of the pose. As it is in 3D, the difference is understood more clearly in the video available at \protect\href{https://youtu.be/KsksVkYRmlg}{https://youtu.be/KsksVkYRmlg}}
    \label{fig:z_axis}
\end{figure}

The fused GPS/INS will often have pose data with frequency over 50 Hz, while the camera will typically operate at a lower frequency. Furthermore, in general, the GPS/INS timestamps will not be synchronised with the camera images as the sensors record the information independently. It is therefore preferable to only compute information necessary for the poses from which the images are captured.
This pose information is computed by the data parser by interpolating the neighbouring poses from the GPS/INS data. The included data parsers employ linear interpolation between the two nearest poses, however, this can be tailored within any given parser.
Poses are colored using a gradient (red to orange by default) making it easy to distinguish poses that belong to overlapping traversals (see \autoref{fig:z_axis}).

\textbf{Performance Optimizations}: The custom parser scripts written in JavaScript are run in a web-client which introduces some language-level processing limitations. In particular, the JavaScript interpreter on a web-client cannot fully utilize all the cores of the CPU for compute-intensive tasks. Also, the use of the GPU for processing in web-clients is limited. 
Some workarounds and alternatives to these shortcomings include running asynchronous operations like fetch and parse in separate worker threads using Web Workers API, and leveraging the client side GPU through libraries such as GLMatrix \cite{glMatrix} to optimize many of the matrix operations used for transforming poses using WebGL operations.
Alternatively, compute intensive operations or GPU-based code can be run on the server. To do this, OdoViz's NodeJS backend can be extended to expose an API endpoint configured to call external code that is natively executed on the server and seamlessly integrated within OdoViz.

\subsection{Offset Settings}
Many datasets and their associated file formats will have their own conventions for coordinate frames e.g. swapped Y and Z axis for positioning and inverted Y rotations. OdoViz allows adjusting the offsets in each of the 3 dimensions for position $P_x$, $P_y$ and $P_z$ and rotation $R_x$, $R_y$ and $R_z$. It is also possible to invert (additively) values assigned to each of the above six fields, swap any two $R$ axes or $P$ axes, and/or scale up or down all axes equally. With known translation and rotation of the camera, it is possible to precisely visualize the vehicle's pose using these offset settings. Further, this conveniently allows loading unadjusted data captured from cameras or INS sensors mounted upside-down, rotated, and/or translated and thus is helpful in making new datasets from captured raw logs. The user interface is carefully designed allowing angle snapping to multiples of 45$^{\circ}$ and providing controls for fine tuning to precisely adjust the scale.

Given the increased error in GPS altitude data \cite{GPS_Altitude_Albri2017} \cite{GPS_Altitude_Garmin} when compared to latitudinal and longitudinal data, loading a journey with the same start and end points can result in a significant error in the z-coordinate (e.g. the Oxford Robotcar 2014-12-10-18-10-50 trajectory results in a significant height difference across the journey). To cater for this, OdoViz allows the altitude to be ignored. Alternatively, the system allows the z-axis to be used to represent time differences i.e. in addition to the trajectory gradient coloring, the z-coordinate increases with time where the rate of increase is adjustable with fine-tuners. This feature is particularly useful when visualizing multiple and overlapping journeys, and in developing extensions and plugins to visualize loop closures such as the HTMap \cite{HTMap_Garcia-Fidalgo2017} extension (explained later in \Cref{section:htmap}).

\subsection{Scene Settings}
By default, the 3D scene is setup with an oblique view, directional light and an auto-expanding grid for reference, with each vehicle pose represented as a low-polygon 3D model of a car. Further settings are provided to switch between different preset view points (e.g. top-view), toggle grid and lights, and to adjust the scale of the elements of the 3D model to cater for different levels of zoom.

Similarly, for data with point-clouds, the size of the points can be adjusted for either the selected pose or the entire dataset. Point-clouds are uniformly colored with perceptually uniform\footnote{if the data goes from 0.1 to 0.2, this should create about the same perceptual change in color as if the data goes from 0.8 to 0.9} and robust to color-blindness \textit{viridis} colormap \cite{Viridis} \cite{Jet_Camgz2003} based on the depth using GLSL shader. User defined colourmaps can also be added. 

Given that the noise associated with point-clouds often increases significantly with depth, these areas consume a disproportionately large region of the color spectrum, leaving a smaller bandwidth for the nearby points.
 To ameliorate this issue, this colour mapping can be adjusted by excluding (farther) points above a certain percentile while coloring. By default we exclude points above the  $90^{th}$ percentile.
OdoViz further provides a setting to add a camera object to each pose, introducing new possibilities for extensions and plugins to add new tasks based on these cameras. For example, we have employed these camera objects to develop a plugin that finds intersecting camera frustums to create datasets of images from overlapping viewpoints.

Mouse interaction to control the viewfinder defaults to Orbit controls (Mouse-drag controls rotation and Alt + Mouse-drag controls translation) however this can be changed to Map controls (or vice versa). Mouse-hover highlights and selects the pose while right mouse-click pins/unpins the currently selected pose. Mouse-hover will not trigger selection when an object is pinned. This is useful when performing other operations on the scene while keeping the desired object selected.

Additionally, there is an animate feature that starts from the first pose and propagates to subsequent poses with an adjustable time delay to smoothly visualized a replay of the entire journey sequentially. As the animation proceeds, poses are selected one after the other, with the viewfinder's target set to the selected object and the selection being pinned to avoid other mouse interactions. This allows rotating and zooming in/out of the map whilst keeping the currently animated object in the center of view.

\begin{figure}
    \begin{subfigure}{0.32\columnwidth}
        \centering
        \includegraphics[width=\columnwidth]{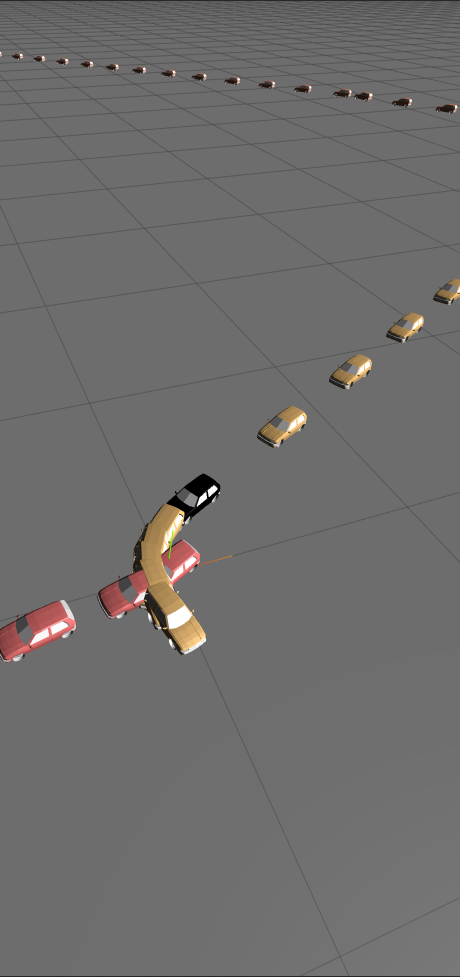}
        \caption{Selected Pose}
        \label{fig:selected_pose}
    \end{subfigure}
    \begin{subfigure}{0.32\columnwidth}
        \centering
        \includegraphics[width=\columnwidth]{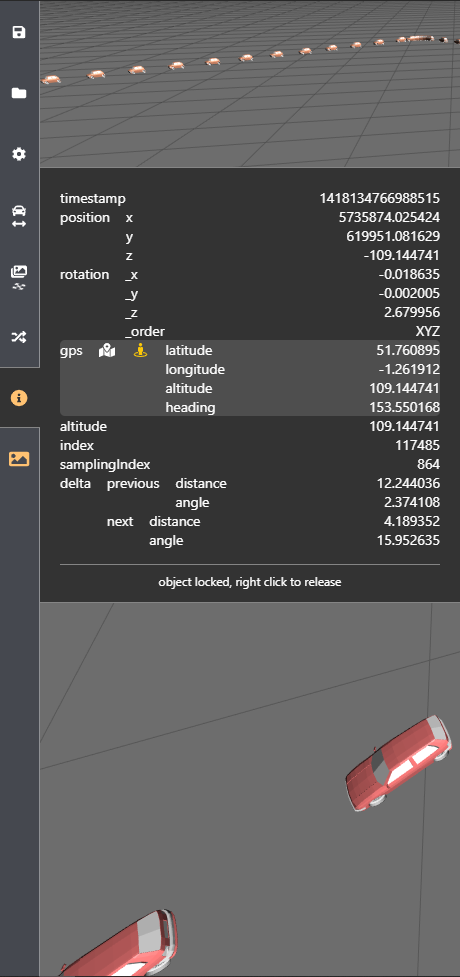}
        \caption{Info Viewer}
        \label{fig:info_viewer}
    \end{subfigure}
    \begin{subfigure}{0.32\columnwidth}
        \centering
        \includegraphics[width=\columnwidth]{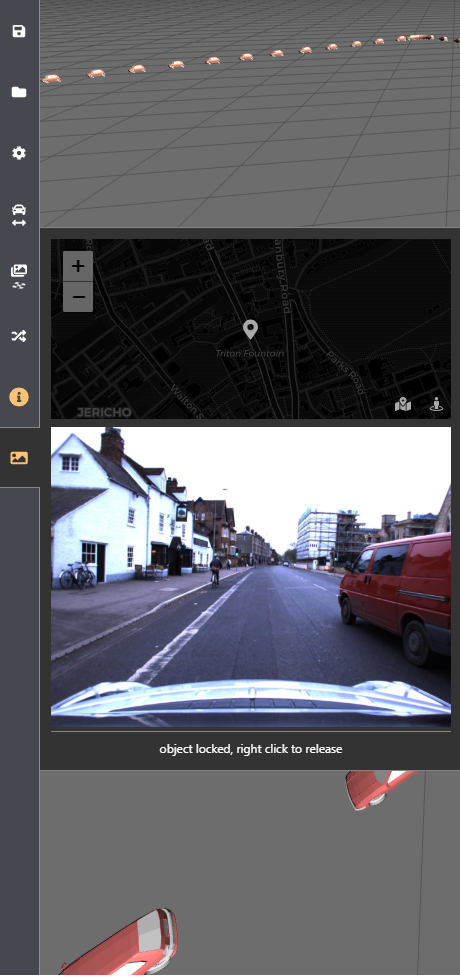}
        \caption{Image Viewer}
        \label{fig:img_viewer}
    \end{subfigure}
    \caption{Screenshots showing (a) the viewport showing the selected pose highlighted in black, (b) information about the selected pose in the Info Panel, and (c) the image along with an embedded minimap in the Image Panel}
\end{figure}

\subsection{Info and Image Viewer}

On selection of a pose in the visualisation tool's main viewport (see \autoref{fig:selected_pose}) 
the associated information and the corresponding image can be viewed in the info panel and the image viewer panel, respectively (i.e. when the panels are activated). This information includes data such as latitude, longitude, altitude, heading, etc., (see \autoref{fig:info_viewer}) and the image taken from the data-point (see \autoref{fig:img_viewer}). We also developed a plugin to conveniently visualize in real-time the selected pose on satellite imagery using its GPS data if available. To do this, we show the selected pose on a mini-map above the image using LeafletJS \cite{leafletjs} and OpenStreetMap \cite{OpenStreetMap} which updates as selection changes. Additionally we integrated a feature that allows viewing the selected pose in a new browser tab on \textit{Google Maps} and on \textit{Google Street-View}. It should be noted that heading information of the current pose is used for comparison of the image against recent 360 captures from the same pose.
\section{Pose Processing}
\label{section:functionality}

\subsection{Adaptive Sampling}


Many datasets offer dense data with poses sampled at a rate that can be as high as 200 Hz. For example, visualizing a 10 km journey through Oxford\footnote{Oxford Robotcar \cite{RobotCarDatasetIJRR} trajectory \href{https://robotcar-dataset.robots.ox.ac.uk/datasets/2014-11-18-13-20-12/}{2014-11-18-13-20-12}} with GPS logged at 50 Hz yields \num{118763} poses whereas images logged at 16 Hz yields \num{35514} poses. Rendering such large number of data-points would require excessive computational resources resulting in very large loading times of up to a few minutes even on a modern high-end consumer-grade CPU. This data is often uniformly sampled using scripts to reduce the amount of data-points rendered in the 3D environment.


This uniform sampling is often also applied when training models that make use of keypoints in the images. In each of these cases, and particularly in the latter, such an approach to sampling may not be suitable. This is due to the fact that a lot of data-points will be decimated around corners despite their exhibiting large variation in the visual content due to the rapid change of heading.

\begin{figure}
    \centering
    \begin{subfigure}{0.24\columnwidth}
        \includegraphics[width=\columnwidth]{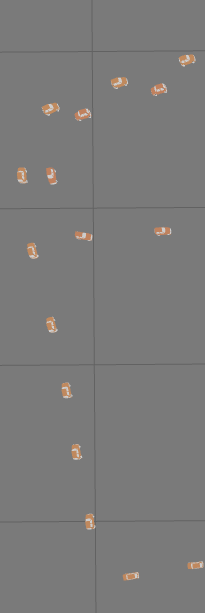}
        \caption{}
        \label{fig:uniform_sampling}
    \end{subfigure}
    \begin{subfigure}{0.24\columnwidth}
        \includegraphics[width=\columnwidth]{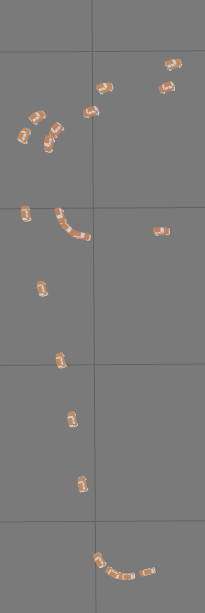}
        \caption{}
        \label{fig:adaptive_sampling}
    \end{subfigure}
    \begin{subfigure}{0.24\columnwidth}
        \includegraphics[width=\columnwidth]{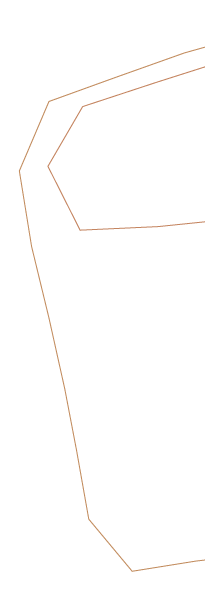}
        \caption{}
        \label{fig:uniform_sampling_contour}
    \end{subfigure}
    \begin{subfigure}{0.24\columnwidth}
        \includegraphics[width=\columnwidth]{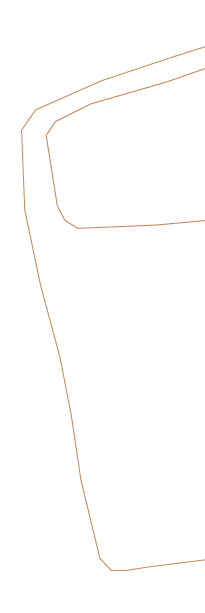}
        \caption{}
        \label{fig:adaptive_sampling_contour}
    \end{subfigure}
    \caption{Screenshots (a) and (b) show poses present in a portion of Oxford Robotcar trajectory after uniform sampling and adaptive sampling respectively. Contours connecting the sampled poses (c) and (d) highlights the difference in results between the two methods and shows how the adaptive sampling preserves the corners of the trajectory. A screen-cast showing how changing the constraints $\tau_{d_\text{acc}}$ and $\tau_{\theta_\text{acc}}$ affects sampling in real-time, is available at \protect\href{https://www.youtube.com/watch?v=9Vf26sgRqSc}{https://www.youtube.com/watch?v=9Vf26sgRqSc}} 
\end{figure}

Usually a sparse set of data points are uniformly sampled from such dense data based on a fixed distance threshold. For example, one pose every 5 metres would reduce the total number of poses from \num{35514} to less than \num{1000} in the above Robotcar trajectory. As such, this distance-based uniform sampling also removes redundant poses at the same GPS co-ordinates captured as the vehicle waits at a red traffic light at an intersection. However, we also lose a lot of visually dissimilar and feature rich images around the corners as the vehicle sweeps a larger heading angle in a very short distance. For example, in a tight turn with 10 or more visually distinct viewpoints, uniform sampling will reduce the resultant segment to one or two images if sampled only based on a distance threshold. 

In order to overcome this issue, we present an adaptive sequence-based sampling technique which is dependent on the rate of change of angle with respect to distance. The algorithm for adaptive sampling is shown in \Cref{alg:adaptive_sampling}. 


\begin{algorithm}
\SetAlgoLined
\KwResult{List of sampled\_poses}
 poses = list of poses\tcp*{population, input}
 sampled\_poses = []\tcp*{result placeholder}
 $d_\text{acc}$ = 0\tcp*{accumulated distance}
 $\theta_\text{acc}$ = 0\tcp*{accumulated angle}
 $\tau_{d_\text{acc}}$ = 12\tcp*{static distance threshold ($m$)}
 $\tau_{\theta_\text{acc}}$ = 15\tcp*{adaptive distance threshold ($\deg$)}
 \ForEach{pose in poses}{
  $d_\text{acc}$ = $d_\text{acc}$ + $\Delta{d}$ \;
  $\theta_\text{acc}$ = $\theta_\text{acc}$ + $\Delta{\theta}$ \;
  \If{$\theta_\text{acc}$ \textgreater \ $\tau_{\theta_\text{acc}}$ or $\theta_\text{acc}$ \textgreater \ $\tau_{\theta_\text{acc}}$}{
    Add pose to sampled\_poses\tcp*{choose sample}
   }
 }
 \caption{Adaptive Sampling}
 \label{alg:adaptive_sampling}
\end{algorithm}

This technique traverses the poses using timestamps and decimates the poses only along the sequence, preserving the poses on overlapping routes within the journey. In contrast binning the poses based on the GPS locations using a KD-Tree and selecting one pose per bin, would yield fewer samples as all poses from a certain GPS location would fall in the same bin regardless of whether it belongs to an overlapping route or not. Hence, the sequence-based sampling is preferred as the resultant set of poses can be used in many place recognition related tasks such as computing loop closure detection on fewer samples from the same trajectory.

A plugin for sampling added to the visualization tool allows us to perform both uniform sampling using a static distance threshold $\tau_{d_\text{acc}}$  and adaptive sampling using an adaptive distance threshold $\tau_{\theta_\text{acc}}$ based on the angle accumulated, $\theta_\text{acc}$, over a distance, $d_\text{acc}$ . We avoid computing interpolation of data as $\tau_{d_\text{acc}}$ or $\tau_{\theta_\text{acc}}$ changes, by pre-computing the interpolated data for all images. This makes it possible to interactively vary both $\tau_{d_\text{acc}}$ and $\tau_{\theta_\text{acc}}$ for adjusting sampling with real-time visual feedback.

\autoref{fig:uniform_sampling} and \autoref{fig:adaptive_sampling} shows the different poses sampled based on the uniform sampling and adaptive sampling, respectively. The plugin also shows total poses that will be sampled based on the criteria chosen and features a facility to export sampled poses as a \textit{json} file for training machine learning models.

\subsection{Finding Pose Correspondences}
\label{sec:finding_pose_corres}
To train computer vision models to extract features from images robust to different viewpoints, seasons and different times of the day, requires many thousands of image pairs from corresponding locations. This data can be curated based on the GPS data. In curating such datasets, it is important to compare images and information regarding the poses of same or another journey traversed on the same route, either partially or completely. We compute a matching pose with the least loss for each of the poses in the journey selected for finding correspondences, where the matching loss between a query pose $x$ and a match candidate pose $y$ is defined as follows:

\begin{equation}
\begin{split}
    \text{Loss} &= \alpha \; \Delta{d} + \beta^{*} \; \Delta{\theta} \\
    \beta^{*} &= \begin{cases}
        \beta \; \dfrac{\theta_\text{acc}}{\tau_{\beta_\theta}}, & \text{if}\ \theta_\text{acc} > \tau_{\beta_\theta} \\
        \beta, & \text{otherwise}
    \end{cases}
\end{split}
\end{equation}
where,
\begin{conditions}
    \alpha & distance importance factor \\
    \beta & angle importance factor \\
    \beta^{*} & adaptive angle importance factor \\
    \Delta{d} & absolute distance\footnotemark \ between $x$ and $y$ \\
    \Delta{\theta} & absolute heading difference between $x$ and $y$ \\    
    \theta_\text{acc} & angle accumulated up to a distance of $\tau_{\beta_d}$ from $y$ \\
    \tau_{\beta_\theta} & beta limiter threshold \\
\end{conditions}
\footnotetext{computed using the Haversine Formula \cite{Haversine_Geodesic_distance_1984S&T....68R.158S} i.e. the great-circle distance between two points on a sphere given their longitudes and latitudes}

In order to speed up the matching process, we discard poses that have a distance loss greater than 30 m and any loss greater than a defined $\tau_\text{loss}$. To avoid having multiple poses matching to more than one query pose, an additional step is performed to check if the match has been paired with any other pose. We update the match only if it has a lower loss than other matches. All the above operations are fully customizable to suit individual needs in a separate \textit{matcher} file.

\begin{figure}
    \begin{subfigure}{0.49\columnwidth}
        \centering
        \includegraphics[width=\columnwidth]{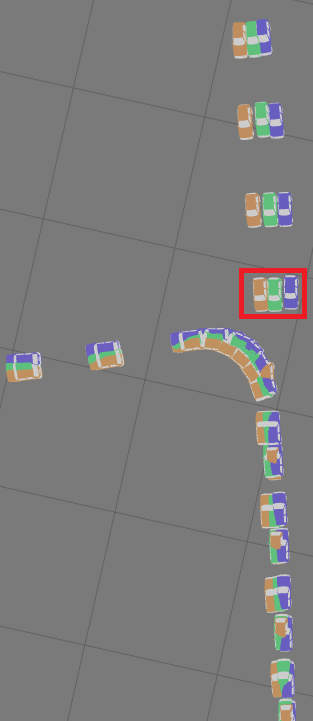}
    \end{subfigure}
    \begin{subfigure}{0.50\columnwidth}
        \centering
        \includegraphics[width=\columnwidth]{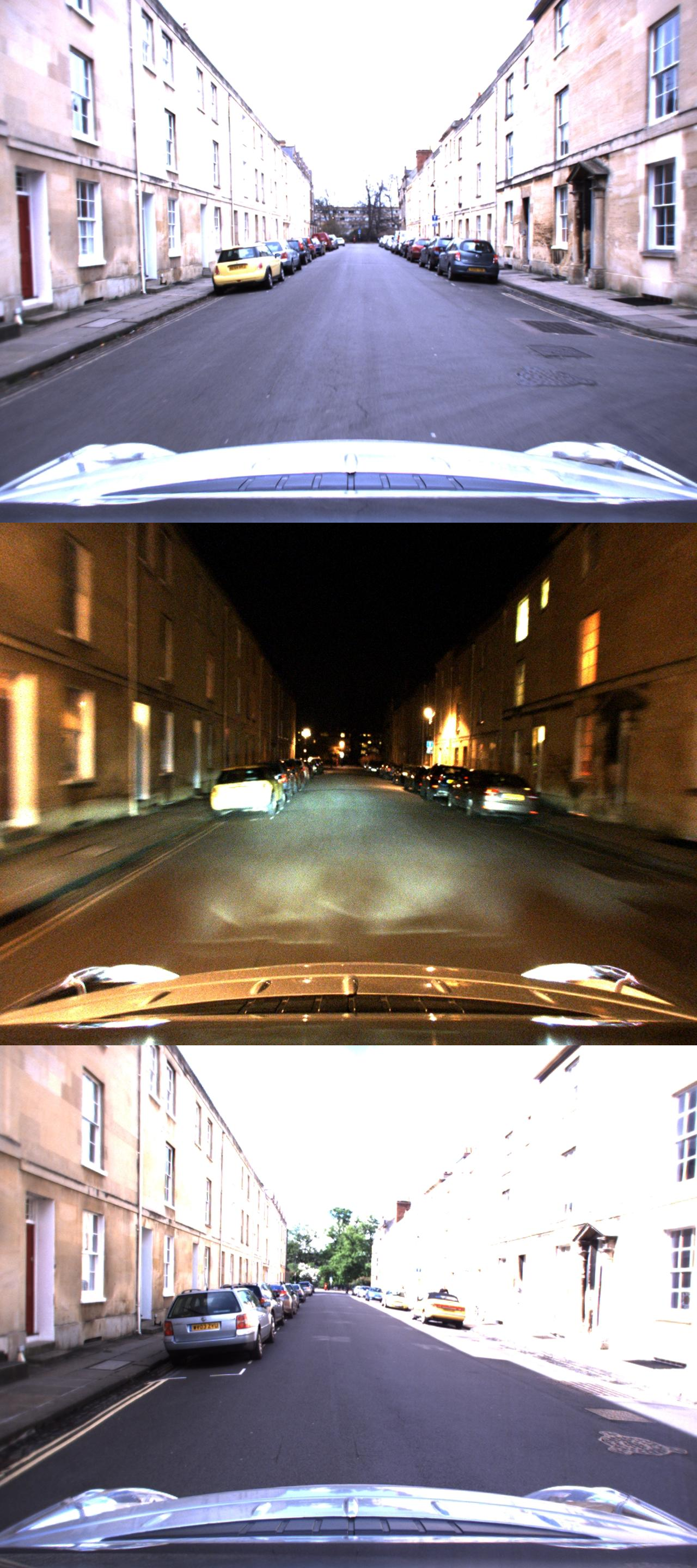}
    \end{subfigure}
    \caption{Left: the main viewport loaded with three Oxford Robotcar trajectories obtained by matching Winter Night 2014-12-10-18-10-50 (green) and Summer Day 2015-05-19-14-06-38 (indigo) against an adaptively sampled Winter Day 2014-12-09-13-21-02 (orange). Right: Images corresponding to the three matched poses marked in red in the left image: top - Winter Day, middle - Winter Night, and bottom - Summer Day.}
    \label{fig:matching}
\end{figure}

When a journey is loaded and sampled, OdoViz can load another overlapping journey and find matching poses for each of the poses in the current journey as described above. The matched poses from the new journey are grouped together, and added to the same 3D scene with a different color. \autoref{fig:matching} shows the color coded matching results of loading two traversals against the loaded traversal. As with the pose sampling plugin, this plugin features an export as \textit{json} option where the resulting file can be used directly for training machine learning models e.g. CNN-based metric learners \cite{metric_learning_contrastive_loss_1640964} \cite{metric_learning_FaceNet_Triplet_Loss_Schroff_2015} \cite{metric_learning_chen_beyond_triplet_loss_quadruplet_loss}.

\section{Extensions and Plugins}
\label{section:extensions_and_plugins}

Further to the core modules described in the previous sections, OdoViz's functionality can be extended through the addition of visualization and compute extensions and plugins for importing and exporting data. Compute intensive extensions are recommended to use Web Workers for compute operations that asynchronously load the compute data. Here we highlight three such extensions that are included with the software in order to provide examples of the versatility of the system. 

\subsection{Save and Restore Plugin}
The Save plugin is a simple import/export plugin that saves store data to the browser's storage. The current view, offset settings, scene settings, sampling settings and the loaded file can be saved and restored using this plugin. Additionally the plugin indicates if there have been any changes made since the last load, for example, to check if the sampling was performed using previously saved settings before exporting sampled poses.

\subsection{top-k Image Retrieval Analysis Extension}
\begin{figure}
    \begin{subfigure}{\columnwidth}
        \centering
        \includegraphics[width=\columnwidth]{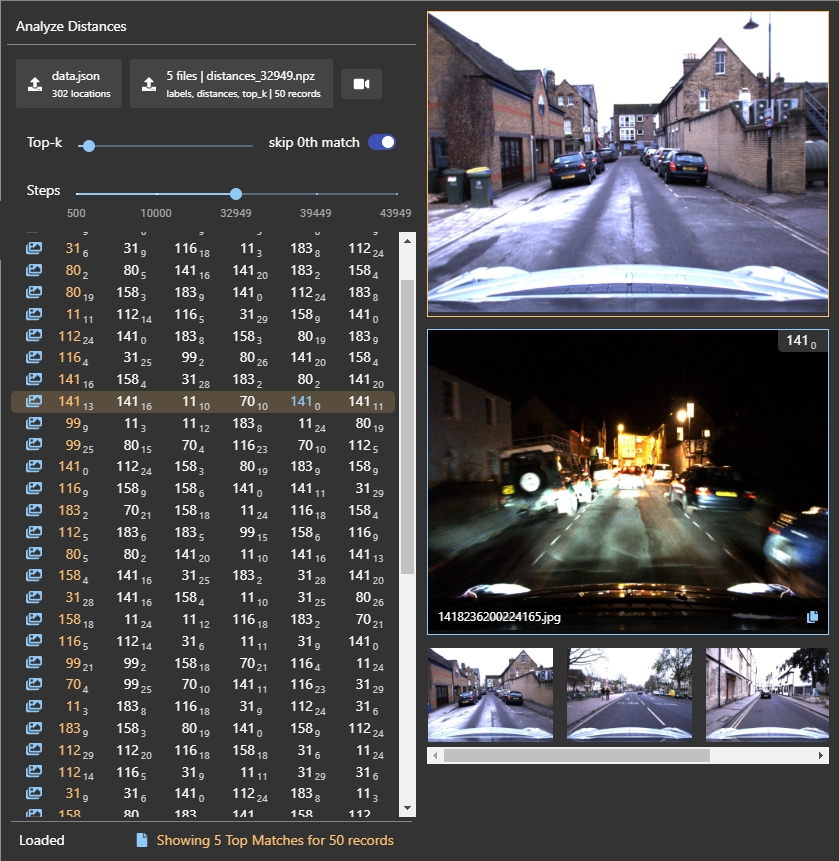}
    \end{subfigure}
    \caption{Screenshot of top-k Image Retrieval Analysis extension showing top-5 matches for all images in a batch at step 32949}
    \label{fig:analyze_distances}
\end{figure}

A common approach for visual place recognition using deep neural networks is to compute a compact embedding that provides a compressed representation of an image's visual appearance suitable for matching and retrieval. When each image can be represented compactly in a low dimensional learned embedding, it is important to visually analyze the precision and robustness of the resulting embeddings. Often a top-k precision and/or top-k recall metric is used to evaluate the quality of the embeddings using retrieval performance. The \textit{top-k Image Retrieval Analysis} extension accepts one or more ($i$) \textit{.npz} file containing pairwise distances, labels and top-k distances output during training for different epochs of a machine learning algorithm and ($ii$) the corresponding \textit{data.json} containing the mapping from label and index to the original image file location on disk. This data is then presented in an intuitive tabulated format showing Top-k matches, with $k=5$ by default. The interface allows the user to interactively explore the results, selecting different query or anchor images, visualising the top-k matches, varying $k$  using a slider, etc. (see \autoref{fig:analyze_distances}). In particular, when a row is selected in the table, an image comparison interface on the right shows images of the ground truth and the top match one below the other, while showing smaller thumbnails of the top-5 matches below. Ground Truth is shown in yellow while the match is shown in blue; as font-color in the table and as border in the image comparison interface. Results for a given epoch can be compared with other epochs using the slider provided.

\subsection{HTMap Extension}
\label{section:htmap}
\begin{figure}
    \begin{subfigure}{0.56\columnwidth}
        \centering
        \includegraphics[width=\columnwidth]{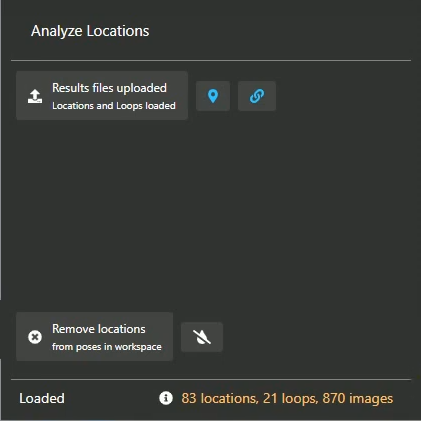}
    \end{subfigure}
    \begin{subfigure}{0.43\columnwidth}
        \centering
        \includegraphics[width=\columnwidth]{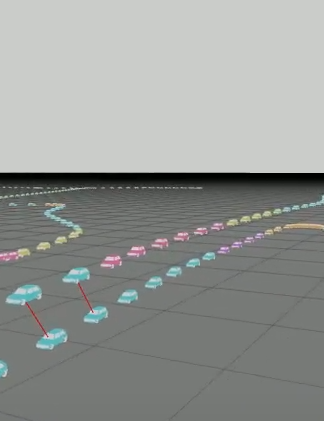}
    \end{subfigure}
    \caption{Screenshot of HTMap plugin showing loop closure connections in an overlapped route within the same trajectory loaded with time in Z-axis}
    \label{fig:htmap}
\end{figure}

As a final example we present a visualisation plugin to render the output of the Hierarchical Topological Maps (HTMap) technique of Garcia et al. \cite{HTMap_Garcia-Fidalgo2017}. This approach divides trajectories by hierarchically grouping images into a set of topologically connected nodes. The HTMap extension allows loading results of HTMap to provide a 3D view of how the trajectory is divided into multiple parts with a different color for each location and where the image loops have been found. \autoref{fig:htmap} shows the HTMap result for one of the Oxford Robotcar trajectories. Notice the different color for sets of poses belonging to the same topological node and the loop closure connections in red that connects the poses of the images matched.

\section{Conclusions}
In the paper we presented {OdoViz}, a new web-tool for simplifying the task of utilising autonomous vehicle datasets within machine learning research. The platform facilitates:

\begin{itemize}
    \item viewing and inspecting complete trajectories supporting many popular datasets out of the box
    \item interactive viewing of information, image and point-cloud data associated with individual poses
    \item viewing poses on satellite imagery inline, externally on Google Maps and the corresponding omnidirectional image on Google Street View.
    \item pose processing including adding offsets, sampling, comparing, and finding pose correspondences within and across the trajectories
    \item exporting resultant poses and correspondences in \emph{json} format suitable for use in machine learning  
    \item adjusting, exporting and importing settings related to the 3D environment and pose processing tasks
    \item adding extensions and plugins to extend the capabilities
    \item integrating server-side compute intensive and GPU-based operations
\end{itemize}

The tool is released as an open-source platform for use and extension by other researchers, and is under further active development within our own group. The tool can be deployed flexibly as single user installation software, browser based tool served from a common organization server, or an online interactive visualization tool for publicly available datasets. Future releases will include:
\begin{itemize}
    \item Rendering pose on standard and custom satellite imagery
    \item Rendering 3D Buildings over satellite imagery with the ability to perform tasks that utilise and operate on the 3D polygon data
    \item A plotting plugin for producing graphs and/or histograms related to the pose data
    \item Supporting and visualising geo-spatial tasks such as clustering, hexbinning, and generating heatmaps from the resultant data
\end{itemize}

\section*{Acknowledgments}

Research presented in this paper was supported by \href{https://www.sfi.ie}{Science Foundation Ireland} grant 13/RC/2094 to \href{https://www.lero.ie}{Lero - the Irish Software Research Centre} and grant 16/RI/3399.

\bibliographystyle{ieeetran}
\bibliography{IEEEabrv, root}

\end{document}